# Improving Neural Network Generalization by Combining Parallel Circuits with Dropout

Kien Tuong Phan[1*], Tomas Henrique Maul[1], Tuong Thuy Vu[1], Lai Weng Kin[2]

[1] University of Nottingham Malaysia Campus, 43500 Semenyih, Selangor, Malaysia,
khyx3pko@nottingham.edu.my
[2] Tunku Abdul Rahman University College, Kuala Lumpur, Malaysia

**Abstract-** In an attempt to solve the lengthy training times of neural networks, we proposed Parallel Circuits (PCs), a biologically inspired architecture. Previous work has shown that this approach fails to maintain generalization performance in spite of achieving sharp speed gains. To address this issue, and motivated by the way Dropout prevents node co-adaption, in this paper, we suggest an improvement by extending Dropout to the PC architecture. The paper provides multiple insights into this combination, including a variety of fusion approaches. Experiments show promising results in which improved error rates are achieved in most cases, whilst maintaining the speed advantage of the PC approach.

Keywords- Deep Learning, Parallel Circuit, Dropout, DropCircuit

## 1 Introduction

Deep learning has repeatedly made significant improvements to the generalization capability of neural networks through several core strategies including: (i) increasing model size, (ii) undergoing longer training periods as well as (iii) adopting larger datasets. Unfortunately, this approach creates a tremendous overhead in terms of both time and computational resources. Deep learning is increasingly becoming more feasible with the support of specialized hardware systems (e.g. multicore CPUs, graphics processing units (GPUs), and high-performance computing (HPC)). However, these solutions tend to be costly and require careful tailoring to derive maximal benefits.

We attempt to apply deep learning to a remote sensing problem, within the constraints of an online platform with limited computational power. In order to address the feasibility concerns mentioned above, we decided to tackle the computational problem at the algorithmic level, as a pure hardware approach would be too expensive to solve the problem alone [1]. Within the scope of our previous paper [2], we have proposed Parallel Circuits (PCs), a biology-inspired Artificial Neural Network (ANN) architecture, as an attempt to reduce heavy computational loads without harming performance. Our preliminary experiments showed that PC architectures could decrease training time up to 40% under some constraints. On the contrary, the impact on classification accuracy was still debatable, as the proposed network exhibited unstable performance across configurations, especially when the size of the original ANN was small.

Dropout is a recent technique for regularization that has been boosting neural network accuracy in many applications. By randomly dropping nodes in the network, dropout successfully prevents the co-adaption of nodes [3]. Thanks to Dropout, over-reliance on specific input nodes is reduced [4]. Taking advantage of this valuable property, in this paper we propose a modification of Dropout whereby we scale it up to the level of Parallel Circuits (i.e. DropCircuit). In this paper, we hypothesize that Dropout would help circuits work more independently (achieving sparser perspectives of the problem domain) and would thus help harvest the benefits of PC modularity. The paper reports on the performance of different types of Dropout-PC combinations.

The paper is structured as follows: sections 2 and 3 provide brief reviews on Parallel Circuits and Dropout respectively; section 4 explains the proposed approach; section 5 describes and discusses the experimental results, comparing several Dropout implementations; and section 6 concludes the paper.

## 2      Parallel Circuits

Our proposed architecture, Parallel Circuits, is inspired from one of nature's solutions to heavy workload, i.e.: parallelism. In implementing PCs, a standard fully-connected Multi-Layer Perceptron (MLP) is divided vertically, forming a series of independent sub-networks called circuits. It's worth mentioning that these circuits share the same input and output layers (i.e. the division is applied only to hidden layers). Thus, compared with MLPs having the same number of nodes, PC architectures reduce the number of connections by a factor of $k$, where $k$ is the number of equal-sized circuits being used. This modification also defines a crucial assumption for PCs that the network should have at least two hidden layers.

It is not unreasonable to expect speed gains from the fact that PCs use fewer connections. On the other hand, it is probably more controversial to hypothesize that PC architectures are advantageous in terms generalization. In [5], an ANN training protocol was reported characterized by problem decomposition (repeatedly switching between multiple goals each with several sub-goals), leading to the emergence of modularity. Provided that modularity is already achieved through the independence of parallel circuits, we reversely hypothesize that this architecture should exhibit the property of automatic problem decomposition. Moreover, and in accordance with the divide and conquer principle, this automatic problem decomposition can further be hypothesized to improve generalization [6].

## 3      Dropout

As mentioned above, Dropout is a regularization technique that has been shown to be effective for a broad variety of neural networks and datasets. Some studies, such as [7,4], have interpreted Dropout as an ensemble method similar to bagging. By randomly dropping nodes, a large number of thinned networks with shared weights are implicitly trained, which when recombined during classification with appropriate scaling, approximate an ensemble of averaged thinned networks [8]. Moreover, in the situation where Dropout is also applied to the input layer, it can be seen as a form of data augmentation whereby noise is added to the input patterns.

Probably due to both its effectiveness and ease of implementation, Dropout is attracting significant attention from researchers in the field. Multiple modifications have been developed for either specialized or general-purpose models. DropConnect is a successful descendent of Dropout, showing even better performance on a range of datasets [9]. In the context of convolutional neural networks (CNNs), one of the most state-of-the-art ANNs for vision-related problems, Dropout has been investigated in different parts of the model. For example, Hinton pioneered the trend of implementing the technique in the final fully connected layers [7], whereas Wu and Xu extended the application of Dropout to pooling layers [3].

## 4 Methodology

### 4.1 Parallel Circuit

According to this definition of Parallel Circuits, for circuit $k$, the input sum for the hidden node $i$ over the layer $l$ might be computed as follows:

$$z_{ki}^l = \sum_j^{n_k^{l-1}} y_{kj}^{l-1} * \omega_{kij}^{l\,l-1} + b_{ki}^l \tag{1}$$

where $n_k^{l-1}$ is the number of hidden nodes on layer $l-1$ of circuit $k$, $y_{kj}^{l-1}$ represents the output of node $j$, in circuit $k$ and layer $l-1$, $\omega_{kij}^{l\,l-1}$ represents the weight of the connection from node $j$ in layer $l-1$, to node $i$ in layer $l$, in circuit $k$, and $b_{ki}^l$ refers to the bias of node $i$ in layer $l$, in circuit $k$. The first hidden layer is actually a special case, as all circuits shared the same input layer ($l = 0$). Therefore,

$$y_{1j}^0 = y_{2j}^0 = \cdots = y_{kj}^0 = x_j \tag{2}$$

On the other hand, the input for the output layer is equivalent to a hidden layer that concatenates the last hidden layers of all circuits (penultimate layer – $PL$), resulting in:

$$z_i^{out} = \sum_k \sum_j^{n_k^{PL}} y_{kj}^{PL} * \omega_{ki}^{out\,PL}{}_j + b_i^{out} \tag{3}$$

### 4.2 Node Dropout

Node Dropout (ND) is an implementation of the standard dropout technique, where *dropping* is applied at the level of nodes. In the PC context, it is worth recalling that we treat each circuit independently. For any single circuit, whenever a training sample is introduced, a distinct mask is generated according to its predefined probability. This favors our attempt to enhance sparsity across subnetworks. Thus, for the PC case, the input sum (1) might be rewritten as follows:

$$m_{kj}^{l-1} = Bernoulli(p_k^{l-1}) \tag{4}$$

$$y'_{kj}^{l-1} = y_{kj}^{l-1} * m_{kj}^{l-1} \tag{5}$$

$$z_{ki}^l = \sum_j^{n_k^{l-1}} y'_{kj}^{l-1} * \omega_{kij}^{l\,l-1} + b_{ki}^l \tag{6}$$

In which, $m_{kj}^{l-1}$ stands for component $j$ of the mask, corresponding to circuit $k$ and layer $l-1$, and the remaining elements are defined as in Equation (1).

### 4.3 DropCircuit

Our PC concept assumes that each circuit should produce a unique perspective of the problem. Therefore, it is critical to prevent circuits from adapting together. When scaling the standard Dropout technique from nodes to circuits, we could consider DropCircuit as a type of "uniform" Node Dropout, where particular nodes stick together for the entire training process. Once one of them is dropped, all the nodes tied to it will also be dropped. To implement this, we create a mask $m_k$ describing the circuit $k$'s dropping status. Every node belonging to the circuit inherits the mask and standard Node Dropout is applied accordingly.

$$m_k = Bernoulli(p) \tag{7}$$

$$y'^{l-1}_{kj} = y^{l-1}_{kj} * m_k \tag{8}$$

In this paper we report on two variants of circuit dropout, i.e.: fixed and non-fixed (mask). In Fixed DropCircuit (FD) we attempt to further specialize circuits by creating a fixed association between sets of patterns and circuits. In this case, a random mask $m_k$ is generated initially for each training instance, and is maintained throughout the training process. On the contrary, Non-Fixed DropCircuit (NFD) works similar to standard Dropout but on the level of circuits, whereby masks are continually regenerated. In other words, Equation (7) will be computed whenever a training sample is introduced, regardless of the epoch.

### 4.4 Experiment Setup

The purpose of the experiments was (i) to prove that Dropout-aided PC can improve generalization performance compared with its counterpart Dropout-aided SC (i.e. Single Circuit - fully connected MLP), and (ii) to determine the most efficient Dropout type for PCs.

The datasets adopted include MNIST and 4 smaller ones obtained from the UCI Machine Learning repository (e.g. Wisconsin Breast Cancer, Breast Tissue, Glass and Leaf) [10]. Two network structures were considered, one with 100 hidden nodes per layer and the other with 1000 nodes; both with two hidden layers. For each network, one SC and 3 PC implementations were examined (corresponding to ND, NFD and FD). The PC versions were further divided into 3 sub categories of 2, 5 and 10 circuits. The models were trained with 0.1 learning rate, 0.4 momentum, 0.0001 L2 sparsity penalty, and the chosen activation function was tanh. In all Dropout-aided conditions, the retaining probability was 0.5. For the four smaller datasets, each condition was run for 100 trials, whereas for MNIST, 10 trials per condition were used. This large number of trials explains our choice of 100 training epochs per training session. The experiments were conducted based on Theano library.

## 5 Experimental results & discussion

### 5.1 Parallel Circuit versus Single Circuit

In this paper, we focus on investigating the ability of PCs fueled with Dropout. Thus, to have a fair comparison, the SC architecture was also implemented with standard

Dropout. Regarding total training times, the problem of PC inefficiencies on small networks is still unsolved. As mentioned in [2] the complexity reduction in small networks is not significant enough to compensate for the computational overhead involved in separating circuits. Therefore, in these networks, PCs, in contrast, increase the total training time to some extent. On the other hand, considering networks with 1000 nodes, we found that implementing PCs gives at least 30% reduction in training time for all cases (Table 2). As mentioned above in Section 1, since one of the successful deep learning strategies is to improve performance by enlarging models, we believe that the reduction is significant. In the small network case, FD is always the approach closest to the speed of SCs while in larger models, both DropCircuit representatives (i.e. FD and NFD) occupy the first and second best positions respectively.

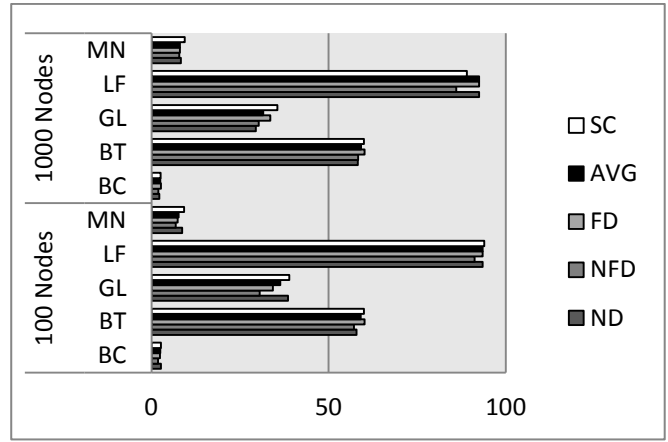

**Fig. 1.** Mean test errors for different conditions and datasets
(MN-MNIST, LF-Leaf, GL-Glass, BT-Breast Tissue, BC-Breast Cancer,
AVG-Average of PCs)

Figure 1 illustrates the test errors for both model sizes, on different datasets, and across Dropout-aided conditions. It's clear that the statistics favor PCs more than SCs with generally lower test errors (39.336% against 40.127%). In fact, PCs always have at least one of its implementations perform better than SCs and its average testing error is lower in most of the cases. Regarding average PC vs. SC performance, this holds true except for only one case in which we used large networks (1000 nodes) to classify the Leaf dataset. For this particular configuration, the average error of all PC implementations (AVG) is higher than that of SCs (92.485% and 89.033% respectively). However, the best individual condition for this case (i.e. 1000 nodes and Leaf dataset) is still a PC condition (i.e. NFD). Significantly, in this context of generalization performance, PCs completely overshadow SCs for small sized networks (100 nodes), contrary to our earlier training time results. Finally, the lowest test errors for each dataset and model size combination were always obtained by a PC architecture.

### 5.2 Node Dropout versus Drop Circuit

From Figure 1, we can see that PCs on average (AVG) obtained a slight advantage over SCs (usually less than 1% in difference). The Glass dataset revealed the largest gaps between SC and AVG, especially with 1000 node networks achieving more than 4% in difference (35.575% versus 31.540% respectively). However, the argument that there is no significant improvement does not hold true. By closer observation, one can see that the performance of the 3 PC-Dropout conditions varied by a large extent. ND, which implements standard Node Dropout, achieved quite similar performance with SC, with a minimal lower average error rate (39.206% versus 40.123%). It exhibited better accuracy than SCs in most of the cases, except for the faulty case (LF + 1000 node family). According to Figure 2, all the median, mean or whiskers of both SC and ND are nearly identical.

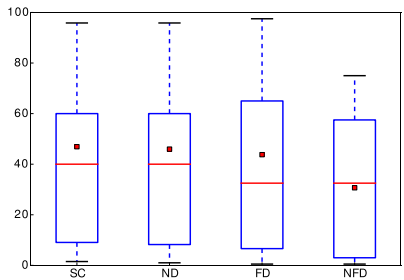

|  |  | SC | ND | NFD | FD |
|---|---|---|---|---|---|
| **100 nodes** | **BC** | 2.710 | 2.655 | 1.842 | 2.375 |
|  | **BT** | 60 | 57.917 | 57.167 | 60.158 |
|  | **GL** | 38.9375 | 38.546 | 30.579 | 34.267 |
|  | **LF** | 93.908 | 93.464 | 91.208 | 93.494 |
|  | **MN** | 9.178 | 8.650 | 6.881 | 7.442 |
| **1000 nodes** | **BC** | 2.575 | 2.235 | 1.917 | 2.668 |
|  | **BT** | 60 | 58.292 | 58.375 | 60.133 |
|  | **GL** | 35.575 | 29.492 | 30.258 | 33.588 |
|  | **LF** | 89.033 | 92.489 | 86.064 | 92.481 |
|  | **MN** | 9.356 | 8.317 | 7.827 | 8.058 |

**Fig. 2.** Average test error of the 4 Dropout-aided conditions

**Table 1.** Testing error of classifiers (dark grey for best, light grey for second best)

NFD and FD are our proposed modification of Dropout for the PC approach (i.e. DropCircuit). NFD, without any hesitation, could be claimed as the champion of the experiments reported here. In every setup (even in the problematic Leaf case), NFD always outperformed SCs (37.212% versus 40.127% on average) and displayed up to 8.5% in error difference for the Glass dataset. The median (32.5%), mean (30.61%) and both whiskers (0.5%, 74.99% respectively) are much lower compared with its counterparts. Recall that in our earlier experiments focusing on training speed [2], the original PC approach was completely beaten by SCs for small scale networks. This time, with NFD, the same setup (small network + PC) achieved the best generalization performance for all 5 datasets. For the Breast Cancer dataset, it scored around 2.375% in error rate, compared with 3.66% in our preliminary test. Thus, we can conclude that NFD is a strong candidate for balancing the speed/generalization trade off in our approach by boosting the accuracy of PCs.

On the other hand, FD only performed better than SC for half of the cases and beat ND 4 times (Table 1). After averaging, FD is still slightly better than SCs (39.466% versus 40.127%) but the median is far lower than that of SC (i.e. the major part in 100 trials achieved better performance compared with SC). One of the possible explanations might be due to the random association between specific data instances and circuits. In future work we will consider versions of FD, which adopt a more informed approach (e.g. via clustering) for associating sample instances with specific circuits. Since FD always achieved the best speed gains and considerably good accuracy, we believe this is a worthy direction for future research.

Finally, apart from providing insights pertaining to different Dropout-aided conditions, the experimental results summarized above, also make it clear that PCs themselves are a useful architecture for improving both speed and generalization. The fact that PCs outperform SCs, for conditions that adopt comparable architectures (e.g.

number of nodes) and training techniques (e.g. Dropout), make it clear that this is an architecture that warrants further investigation in the context of Machine Learning and Neural Computation. Moreover, the results possibly shed additional light onto the rationale for parallel circuits in biological neural networks (BNNs). It is quite possible that parallel circuits in BNNs are partly motivated by the implementation of implicit ensembles aided by neurophysiological mechanisms related to Dropout.

### 5.3 MNIST

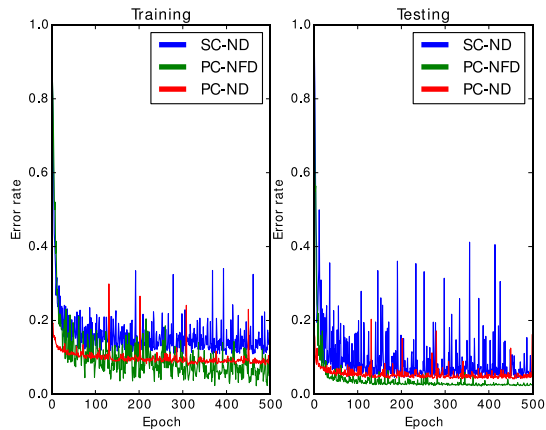

**Fig. 3.** Training and Testing error across epochs

In Section 5.2, we showed that the PC approach could reduce training times. To test whether this result scaled to larger datasets, we compared conditions using MNIST, but this time extended training to 500 epochs using a [1000-500] network structure with a much higher learning rate (i.e. 1) to favor Dropout. Other parameters were kept unchanged. Figure 3 points out that both of the PC approaches converge faster, reaching a plateau around epoch 20 (especially PC-ND) compared to epoch 60 of SCs. Significantly, in this period, PC approaches reached lower plateaus than that of SC. Both of the PC versions exhibited the normal pattern of Dropout implementations, with error fluctuations throughout training. Especially in PC-NFD, the degree of fluctuation was even higher than PC-ND. In the end, PC-ND achieved slightly lower error rate than SC (best at 4.19% and 4.38% respectively) while PC-NFD reached 2.22%. This revealed that (i) with smaller epoch numbers, PC approaches can achieve test errors at least similar to SCs and (ii) PC-NFD performs much better than NDs.

## 6 Conclusion

In this paper, we proposed an improvement of the parallel circuit approach by implementing Dropout, specifically targeting generalization performance. The experiments showed that combining Parallel Circuits with Dropout not only reduces training times but also enhances generalization performance in most cases. Our work provides multiple insights pertaining to this combination, and includes a benchmark comparing different variants. We found that Non-Fixed DropCircuit leads to the best improvements in generalization performance.

# Appendix

|  |  | SC | ND | NFD | FD |
|---|---|---|---|---|---|
| **100 nodes** | BC | 2.293 | 4.723 | 4.957 | 2.819 |
|  | BT | 0.386 | 0.826 | 0.857 | 0.506 |
|  | GL | 0.814 | 1.707 | 1.766 | 1.037 |
|  | LF | 1.490 | 2.938 | 3.111 | 1.794 |
|  | MN | 362.914 | 424.498 | 412.547 | 372.583 |
| **1000 nodes** | BC | 89.073 | 29.930 | 29.606 | 25.557 |
|  | BT | 16.605 | 5.258 | 5.020 | 4.399 |
|  | GL | 34.542 | 10.539 | 10.346 | 8.865 |
|  | LF | 53.541 | 18.392 | 18.316 | 15.796 |
|  | MN | 6475.711 | 4060.208 | 3958.506 | 3595.458 |

**Table 2.** Training time of classifiers (dark grey for best, light grey for second best)